\documentclass[runningheads]{llncs}
\usepackage[T1]{fontenc}
\usepackage{enumerate}
\usepackage[inline]{enumitem}
\usepackage{graphicx}
\usepackage{adjustbox}
\usepackage[utf8]{inputenc}
\usepackage{booktabs}
\usepackage{makecell}
\usepackage{xcolor}
\usepackage{amsmath}
\usepackage{subcaption}  
\usepackage{amsfonts}
\usepackage{url}
\usepackage{multirow}      
\usepackage{caption}       
\usepackage{array}    

\begin{document}
\title{A Computer-aided Framework for Detecting Osteosarcoma in Computed Tomography Scans}
\titlerunning{Computer-aided Framework for Detecting Osteosarcoma in CT Scans}
%
\author{Maximo Rodriguez-Herrero\inst{1} \and
Dante D. Sanchez-Gallegos\inst{1}\orcidID{0000-0003-0944-9341} \and Marco Antonio N\'u\~{n}ez-Gaona\inst{2} \and Heriberto Aguirre-Meneses\inst{2} \and  Luis Alberto Villalvazo Gutiérrez\inst{2} \and Mario Ibrahin Gutiérrez Velasco\inst{3}  \and J.L. Gonzalez-Compean\inst{4}\orcidID{0000-0002-2160-4407} \and Jesus Carretero\inst{1}\orcidID{0000-0002-1413-4793}}
\authorrunning{Rodriguez et al.}
%
\institute{Universidad Carlos III de Madrid, Departamento de Informática, ARCOS, Avenida de la Universidad, 30 (edificio Sabatini), 28911 Leganés (Madrid), España \\ \email{maxrodri@inf.uc3m.es,dantsanc@pa.uc3m.es,jesus.carretero@uc3m.es} \and
Instituto Nacional de Rehabilitacion ``Luis Guillermo Ibarra Ibarra'', Mexico City, México \\ \and
SECIHTI - Instituto Nacional de Rehabilitacion ``Luis Guillermo Ibarra Ibarra'', Mexico City, México \\ \and
Cinvestav Tamaulipas, Cd. Victoria, Tamaulipas, México \\
\email{joseluis.gonzales@cinvestav.mx}}
\maketitle              
\begin{abstract}
Osteosarcoma is the most common primary bone cancer, mainly affecting the youngest and oldest populations. Its detection at early stages is crucial to reduce the probability of developing bone metastasis. In this context, accurate and fast diagnosis is essential to help physicians during the prognosis process. The research goal is to automate the diagnosis of osteosarcoma through a pipeline that includes the preprocessing, detection, postprocessing, and visualization of computed tomography (CT) scans. Thus, this paper presents a machine learning and visualization framework for classifying CT scans using different convolutional neural network (CNN) models. Preprocessing includes data augmentation and identification of the region of interest in scans. Post-processing includes data visualization to render a 3D bone model that highlights the affected area.  An evaluation on 12 patients revealed the effectiveness of our framework, obtaining an area under the curve (AUC) of 94.8\% and a specificity of 94.6\%.

\end{abstract}

\section{Introduction}
\label{sec:introduction}

Osteosarcoma is the most common primary bone cancer. 
Despite advances in treatment, the prognosis for osteosarcoma has not significantly improved in recent years, highlighting the need for continued research, particularly in the area of early detection. 
In recent years, machine learning has emerged as an alternative for analyzing medical images such as  X-rays, magnetic resonance imaging (MRI), and computed tomography (CT) scans to generate outputs such as tumor detection~\cite{Yu2021ConvolutionalPerspectives}.

A major challenge is the shortage of trained specialists available to perform image analysis~\cite{Kmietowicz2017RadiologistCollege}, a task that is not only time-consuming but also increasingly in demand as cancer diagnoses continue to rise. Hence, developing machine learning tools capable of providing automatic diagnoses, such as tumor detection and segmentation, can significantly reduce the workload of radiologists and radiation oncologists. Such assistance enables healthcare systems to serve more patients efficiently, while ensuring that each receives more personalized and attentive care.

In this context, this paper presents an assisting framework for clinicians that automates osteosarcoma detection through a pipeline for processing CT images, including pre-processing, classification, post-processing, and 3D visualization of osteosarcoma. In the pre-processing stage, we identify the region of interest (ROI) in CT images, focusing on the area surrounding the bone and removing external objects such as the scanner table. We also apply data augmentation techniques to promote generalization. The pre-processed data is forwarded to the classification stage, where we employ state-of-the-art convolutional neural network (CNN) models to detect tumors in CT images. Specifically, we evaluated models from the ResNet (ResNet18 and ResNet50)~\cite{He2016DeepRecognition} and DenseNet (DenseNet121 and DenseNet169)~\cite{HuangDenselyNetworks} architectures. We utilized \textit{transfer learning} during the final training setup to address the limited availability of labeled data. To aid physicians in both diagnosis and prognosis, the final stages of our framework (post-processing and visualization) generate 3D bone model visualizations that highlight tumors in positively classified images.

We developed and validated our framework using CT scans provided and labeled by the Mexican National Rehabilitation Institute "Luis Guillermo Ibarra Ibarra" (INR LGII). This dataset includes bounding-box labeled images that were previously anonymized at the institute. We report multiple evaluation metrics obtained using this dataset, including 94.8\% area under the curve and specificity of 94.6\%.

In summary, the main contributions of this paper are:
\begin{enumerate*}[label=(\textit{\roman*)}]
    \item The design of a generalizable framework that facilitates the detection of osteosarcoma in CT scans.
    \item A fully automatic pipeline to preprocess, classify, post-process, and visualize large bone CT scans.
    \item The training and evaluation of multiple CNN models using a dataset acquired at a Mexican healthcare institution.
\end{enumerate*}

The rest of the paper is organized as follows: \S~\ref{sec:related-work} reviews related work. \S~\ref{sec:methodology} describes the methodology used to design the proposed framework. \S~\ref{sec:experiments} presents the evaluation of different models for osteosarcoma classification in CT scans. Finally, \S~\ref{sec:conclusions} provides conclusions and outlines future directions.


\section{Background and Related Work}
\label{sec:related-work}

One major issue in computer-aided systems using deep learning is the limited availability of annotated data. 
Medical datasets are typically much smaller,
and comprehensive annotations are often incomplete or entirely unavailable. Another critical challenge is the high standard of reliability required in medical applications. 

Several studies in the literature have explored the detection of osteosarcoma using machine learning. Mahore et al. \cite{mahore2021machine} used a Random Forest classifier on histopathological images obtained from biopsy samples. While this approach provides high-resolution cellular detail, it relies on invasive procedures that are not always feasible, particularly in low- and middle-income countries. In contrast, medical imaging modalities such as computed tomography (CT) scans offer a non-invasive alternative. CT imaging is especially suitable for studying osteosarcoma due to its high resolution and ability to capture detailed bone structures.

In this context, Sampath et al. \cite{sampath2024comparative} present a comparative study of various deep learning architectures for bone cancer detection. Their pipeline includes preprocessing with a median filter, followed by K-means clustering and edge detection for bone segmentation. They report a testing accuracy of 100\% using an AlexNet CNN on a dataset of 1,141 CT scan images. While these results are promising, such a high level of accuracy may indicate overfitting or dataset leakage, especially given the modest dataset size.

In this paper, we not only focus on cancer detection in CT scans, but also on the creation of 3D visual models to aid physicians in interpreting the results and providing accurate diagnoses.  Our approach builds upon the pipeline proposed by Primakov et al. \cite{Primakov2022AutomatedImages}, originally developed for lung tumor analysis. We adapt and extend their methodology for the detection and visualization of osteosarcoma. 

To mitigate the challenge of limited labeled data, we rely on unsupervised segmentation techniques combined with supervised classification models to localize affected regions. Additionally, unlike the method in \cite{Primakov2022AutomatedImages}, which trains models from scratch, our approach leverages state-of-the-art transfer learning techniques \cite{Kim2022TransferReview}. This allows us to make better use of limited data while achieving high model performance.


\section{Methodology}
\label{sec:methodology}

This section describes the methodology used to design a framework for automatically detecting osteosarcoma using CNN models. First, we describe a general overview of this framework and the steps used to design it, including the preparation of the dataset to train and evaluate different CNN models. 

\begin{figure}
    \centering
    \includegraphics[width=0.6\linewidth]{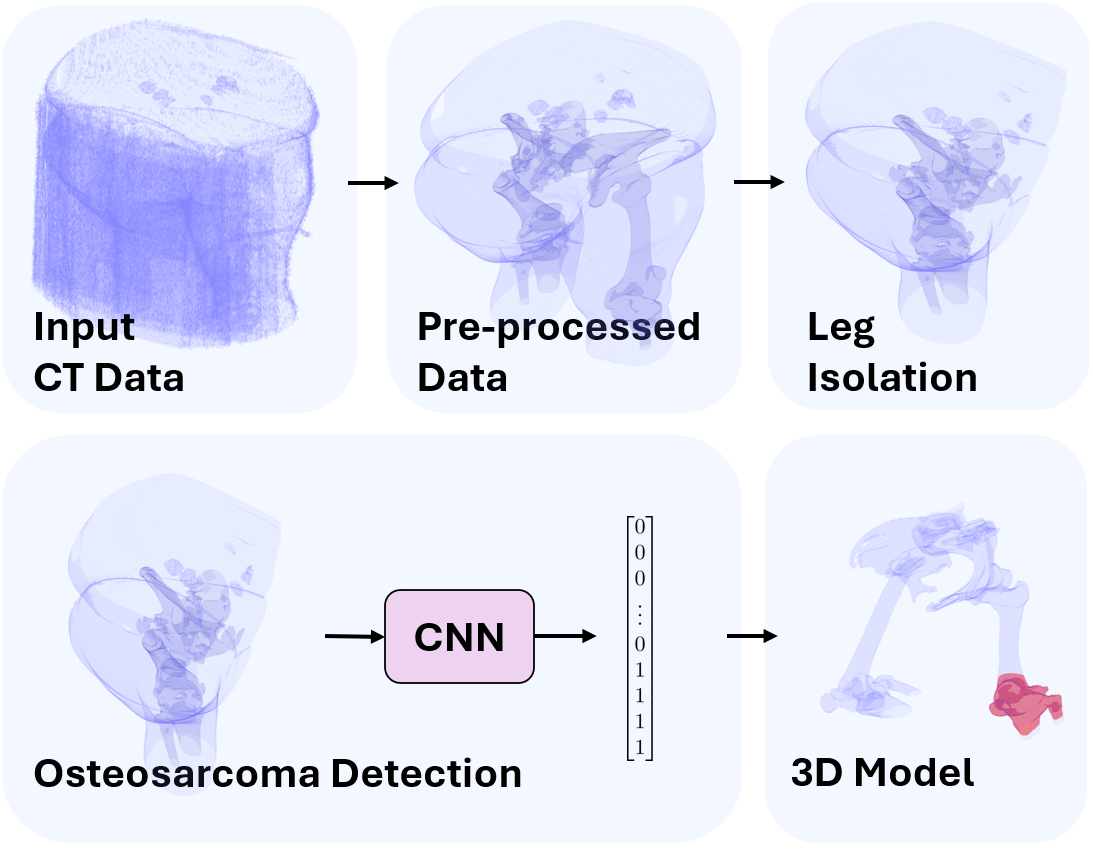}
    \caption{Proposed framework for identifying osteosarcoma using CT images.}
    \label{fig:proposedsystem}
    \vspace{-0.5cm}
\end{figure}

\subsection{Proposed Framework}

Figure~\ref{fig:proposedsystem} provides an overview of the proposed framework for automating the identification of osteosarcoma. The process begins with CT scans, which are passed as input to the framework. These images are first pre-processed to ensure consistency with the data used to train the CNN model. Subsequently, the images are refined further to isolate the bones of the leg (primary region of interest). These preprocessing steps enhance the robustness of the CNN used in the tumor detection stage. Finally, the output of the CNN is utilized to generate a 3D visualization model, supporting the prognosis of osteosarcoma.

\subsection{Dataset Construction}
\label{sec:dataset}
 
The classification dataset consists of 2,910 CT slices from 57 patients. The images were sourced from the PACS-INR (Picture Archiving and Communication System) of the Instituto Nacional de Rehabilitación "Luis Guillermo Ibarra Ibarra" \cite{nunez2024product}. Patients were aged between 20 and 50 years, skeletally mature, and had a confirmed diagnosis of femoral osteosarcoma, with CT studies acquired between 2015 and 2022. All CT images in DICOM format were anonymized by removing identifiable metadata, such as patient and physician information. Axial images were further processed to generate sagittal and coronal anatomical planes, and converted to RGB grayscale at a resolution of 512$\times$512 pixels (16-bit depth) for consistency. 

Of these, 45 patients, consisting of $2,284$ images (78.49\%), were allocated to training and validation, and 12 patients, $626$ images (21.51\%), were reserved as an independent test set. Careful attention was given to ensure that the test set included both patients with few slices and those with more extensive scans, achieving a diverse and representative evaluation subset. Importantly, all dataset splits were performed at the \textit{patient level}. Splitting data at the image level can lead to nearly identical slices from the same patient appearing in the training, validation, and test sets. This compromises the integrity of reported metrics and misrepresents the model’s true generalization capability. Patient-wise splitting ensures a more rigorous and reliable evaluation of the model.

\subsubsection{Data Pre-processing}
The CT images in the dataset were pre-processed using various techniques to eliminate external artifacts (e.g., the scanner table) that could negatively impact the performance of CNN models. Additionally, the leg section was centered within a $256\times256$ pixel region of interest (ROI). This pre-processing step was designed to enhance the dataset’s robustness by consistently defining the ROI, thereby improving the performance and reliability of CNN models.

\begin{figure}
    \centering
    \includegraphics[width=1\linewidth]{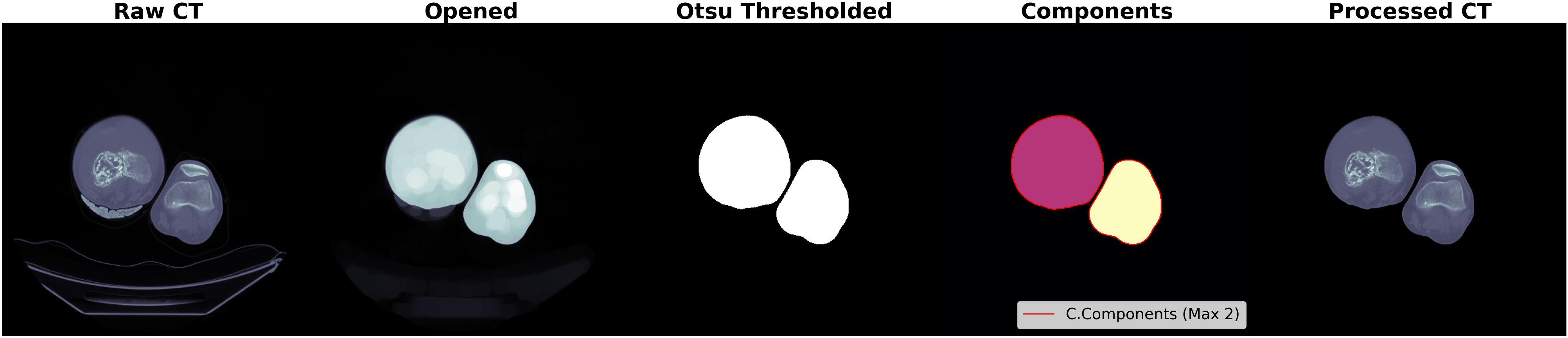}
    \caption{Automated pre-processing pipeline applied to each patient's CT scan. From left to right: raw CT image with scanning table artifacts, result of morphological opening using a disk-shaped structuring element (radius = 10px), binary mask obtained via Otsu thresholding, and final selection of the two largest connected components representing the leg region.}
    \label{fig:preprocessing}
    \vspace{-0.5cm}
\end{figure}

Figure~\ref{fig:preprocessing} shows the preprocessing pipeline used to get the ROI masks.
The process begins by loading each scan, followed by a morphological opening operation using a disk-shaped structuring element with a radius of 10 pixels. This operation erodes soft element connections and smooths out smaller structures, effectively removing or separating the table from the patient anatomy. Next, Otsu's unsupervised thresholding method \cite{otsu} is applied to generate a binary mask that accurately segments the foreground. From this mask, the two largest connected components at the top of the scan are retained. Finally, the original 512$\times$512 images are cropped into two primary ROIs representing each leg, then intensity-normalized slice-wise and resized to 256$\times$256 pixels using nearest-neighbor interpolation. 

\subsubsection{Embedding-Based Misleading Label Detection}

To address annotation inconsistencies in the dataset, specifically cases where visually similar images were assigned conflicting labels, we implemented an embedding-based analysis pipeline using a pretrained ImageNet model. Features were extracted by forwarding image batches through the model’s backbone (classification head removed), followed by global average pooling to produce fixed-length embedding vectors. These embeddings capture high-level semantic content, enabling meaningful comparisons between images.

We computed pairwise cosine similarity across all embeddings and flagged image pairs with high similarity (threshold > 0.95) but differing labels as potential mistakes. This process identified 144 suspicious image pairs. Given the clinical priority of tumor detection, we conservatively removed the images labeled as healthy from these pairs. This decision helps minimize false negatives, which is crucial in clinical settings. In total, 72 images were excluded, resulting in a curated dataset of 2,838 images. The test split remained unaltered to ensure consistent and unbiased evaluation, even as the rest of the dataset underwent curation.

\subsubsection{Data Augmentations}
Given the limited size of the available dataset, a carefully selected set of data augmentations was applied to promote generalization and reduce the risk of overfitting. The augmentation pipeline includes random flips along one or both axes ($p = 0.5$), small rotations ($\theta \in [0, \frac{\pi}{12}]$, $p = 0.2$), random zooming (scaling factor in the range $[0.95, 1.05]$, $p = 0.2$), Gaussian noise injection ($\mu = 0.0$, $\sigma = 0.05$, $p = 0.2$), intensity shifting (offset of $0.2$, $p = 0.8$), and coarse dropout (randomly occluding 5 to 10 patches of size 16$\times$16 pixels, $p = 0.2$). All transformations were implemented using the MONAI framework~\cite{JorgeCardoso2022MONAI:Healthcare}.

\subsection{Tumor Classification }
\label{sec:tumor-classification}


\subsubsection{Base CNN Determination}

For the detection of osteosarcoma in CT images, we implemented and evaluated several CNN models from the ResNet and DenseNet families,  widely adopted in medical imaging applications, particularly under limited data conditions~\cite{Chen2022RecentAnalysis,Litjens2017AAnalysis,Kmietowicz2017RadiologistCollege}.

From the DenseNet family, we selected the DenseNet121 and DenseNet169 variants, while from the ResNet family, we chose ResNet18 and ResNet50. To assess model suitability, all networks were trained from scratch under identical conditions, with validation performed after each epoch. Training was conducted over 15 epochs using a batch size of 16, and the model achieving the highest validation performance across all epochs was retained.

The training process utilized cross-entropy loss~\cite{zhang2018generalized} and the Adam optimizer with a learning rate of $10^{-5}$. Model outputs were passed through a softmax activation function prior to loss computation.





\subsubsection{Transfer Learning}
Transfer learning (TL) has become a standard practice in medical imaging and is frequently employed in state-of-the-art approaches to address the scarcity of annotated data~\cite{Chen2022RecentAnalysis,Liu2021AMethods}. By initializing models with weights pre-trained on large-scale image datasets, such as ImageNet~\cite{Deng2009ImageNet:Database}, it is possible to transfer and adapt learned representations to the target domain through fine-tuning. In this work, we compare three widely adopted fine-tuning strategies~\cite{Davila2024ComparisonClassification}, all initialized using ImageNet weights\footnote{\url{https://pytorch.org/vision/main/models.html}}. 

Each strategy begins by loading the model with pre-trained weights. The strategies are as follows:
\begin{enumerate*}[label=(\textit{\roman*)}]
    \item \textbf{Full Fine-Tuning (FT)}: All model layers are fine-tuned from the beginning.
    \item \textbf{Linear Probing followed by Fine-Tuning (LP-FT)}: Training begins with only the classification head, keeping the rest of the model frozen. When the validation area under the curve plateaus for five epochs, all layers are unfrozen, and training resumes in the full model.
    \item \textbf{Gradual Layer-wise Fine-Tuning (G-LF)}: Training starts with the classification head, and progressively unfreezes one layer at a time, starting from the last, whenever the validation area under the curve stagnates for five epochs. This process continues until all layers are unfrozen or early stopping occurs.
\end{enumerate*}

For TL, we extended the number of epochs to 75, which allows sufficient adaptation during fine-tuning, with early stopping in place to prevent overfitting. The optimizer used across all strategies was Adam. For FT, we used a cosine annealing scheduler with a learning rate range of $[10^{-5}, 10^{-6}]$ and $T_{\text{max}} = 50$, with patience of 5 epochs for early stopping. For LP-FT, the classification head was trained using a \texttt{ReduceLROnPlateau} scheduler, triggered by stagnation in validation AUC. If no improvement was observed for 3 consecutive epochs, the learning rate was halved down to a minimum of $10^{-6}$. Once full fine-tuning began, the same cosine annealing scheduler as FT was applied. For G-LF, the process mirrored LP-FT, but layers were unfrozen one at a time after 5 epochs of AUC stagnation, until all layers were active or early stopping conditions were met.

\subsection{Generation of 3D Models for Data Visualization}
\label{sec:data-visualization}

\subsubsection{3D Bone Model}
\begin{figure}[ht]
    \centering
    \includegraphics[width=1\linewidth]{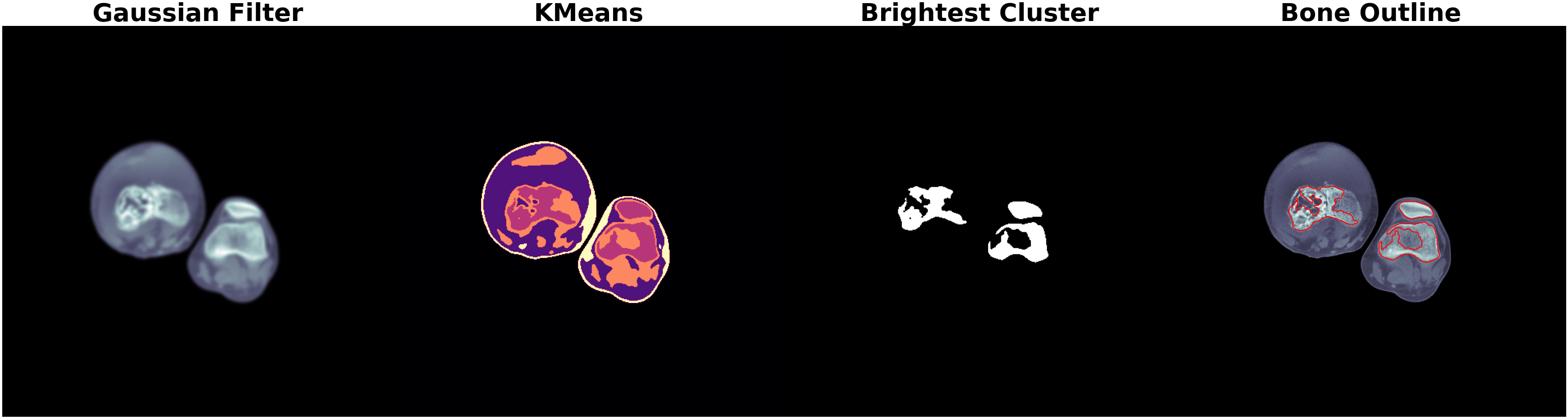}
    \caption{Slice-wise bone extraction pipeline applied to pre-processed CT images. From left to right: Gaussian smoothing ($\sigma=2$), K-means clustering with $k=5$, selection of the brightest cluster to isolate bone tissue, and final binary mask after morphological closing and hole filling outlined on the pre-processed CT scan.}
    \label{fig:bone-mask}
\end{figure}

A series of image post-processing steps is applied to extract the bone structures from CT images. Figure~\ref{fig:bone-mask} depicts these steps. First, a Gaussian filter with $\sigma=2$ is applied to reduce noise. Next, K-means clustering is performed with $k=5$ to segment the image intensities, retaining only the brightest cluster to isolate bone structures. This is followed by morphological closing and binary hole filling operations, applied slice-wise.

Subsequently, the 3D model is generated from the identified bone structures. A volumetric morphological closing using a spherical structuring element with a radius of 3 pixels is applied to enhance connectivity between each bone slice. This includes removing floating artifacts smaller than a configurable threshold. 
The resulting 3D model is then smoothed using Taubin smoothing~\cite{taubin1995curve}, as implemented in the PyVista library~\cite{sullivan2019pyvista}.


\subsubsection{Affected Region Post-processing}

\begin{figure}
\vspace{-1.5cm}
    \centering
    \begin{subfigure}[b]{0.49\textwidth}
        \centering
        \includegraphics[width=1.2\linewidth]{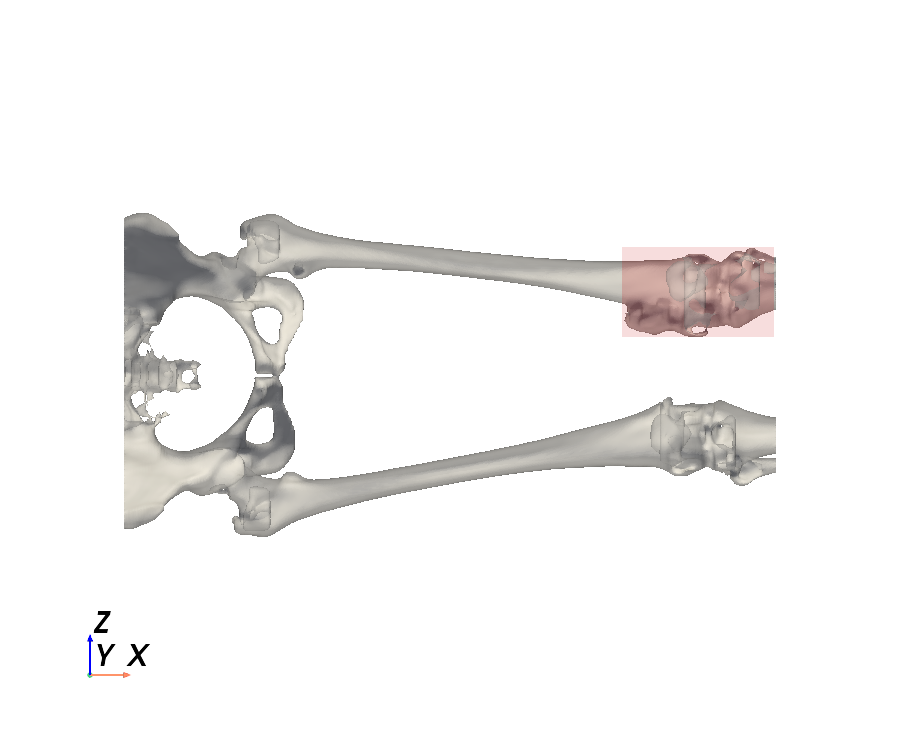}
    \end{subfigure}
    \hfill
    \begin{subfigure}[b]{0.49\textwidth}
        \centering
        \includegraphics[width=1.2\linewidth]{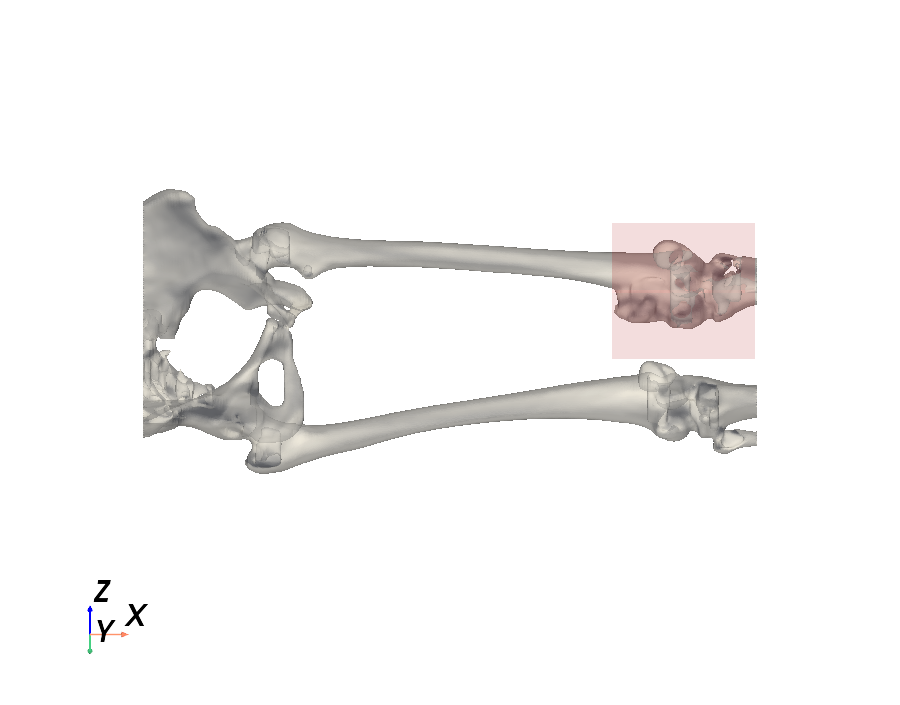}
    \end{subfigure}
    
    \caption{3D model output of the framework for an osteosarcoma-positive patient, shown from two different viewing angles. The red bounding box delineates the predicted affected region.}
    \label{fig:render-result}
\vspace{-1cm}
    
\end{figure}

As shown in Figure~\ref{fig:render-result}, in a 3D model, the area with tumor is delimited using a 3D bounding box to highlight the affected area visually. This visualization aids physicians in rapidly localizing the suspected tumor region, facilitating clinical interpretation and decision-making. 

This region is obtained by passing each slice to the CNN model to identify those classified as tumor. A confidence threshold of 0.95 is used to classify slices as osteosarcoma-positive. We identify the bounding slice indices that delineate the predicted tumor region from the resulting high-confidence regions in the prediction array. Furthermore, a Gaussian smoothing filter with $\sigma=2$ is applied to the confidence array to reduce prediction noise. Additionally, a median filter with a kernel size of 3 is used to suppress spatial inconsistencies and abrupt transitions between tumor and non-tumor slices, thereby promoting spatial continuity.


\section{Experiments and Results}
\label{sec:experiments}
\subsubsection{Evaluation Metrics}

We used the Area Under the Receiver Operating Characteristic Curve (ROC AUC) as a global performance measure to evaluate tumor detection performance. In addition, we report class-specific metrics to provide a more detailed understanding of the model’s behavior.

\paragraph{\textbf{Sensitivity}} (also known as the true positive rate) quantifies the proportion of actual positive samples that are correctly identified by the model:
\begin{equation}
    \text{Sensitivity} = \text{TPR} = \frac{\text{TP}}{\text{TP} + \text{FN}}
\end{equation}

\paragraph{\textbf{Specificity}} (also referred to as the true negative rate) measures the proportion of actual negative samples that are correctly classified:
\begin{equation}
    \text{Specificity} = \text{TNR} = \frac{\text{TN}}{\text{TN} + \text{FP}}
\end{equation}

Here, TP, FP, TN, and FN denote the number of true positives, false positives, true negatives, and false negatives, respectively.

All the reported metrics represent the mean over five cross-validation folds, with 95\% confidence intervals computed using Student's $t$-distribution.

\subsubsection{Baseline Determination}
First, we evaluate different CNN models from the DenseNet and ResNet families. This experiment aims to choose the model with the best performance according to the metrics above. Table~\ref{tab:baseline-model-performance} summarizes the results obtained across five folds. As can be observed, the DenseNet family stands out as the most suitable architecture for the current task.
Furthermore, there was no statistically significant difference in performance between DenseNet121 and DenseNet169, so we selected DenseNet121 as the baseline model. This choice balances performance with computational efficiency and helps mitigate the risk of overfitting, which is a concern when using larger models on limited datasets. 

\begin{table}
\vspace{-0.5cm}
\centering
\caption{Baseline models performance metrics (mean ± 95\% CI) across 5 folds.}
\renewcommand{\arraystretch}{1.2}
\setlength{\tabcolsep}{8pt}
\begin{adjustbox}{width=\linewidth}
\begin{tabular}{lccc}
\toprule
\textbf{Model} & \textbf{AUC} & \textbf{TPR} & \textbf{TNR} \\
\midrule

DenseNet121
& 0.828 (0.765--0.892)
& \textbf{0.671} (0.502--0.840)
& \textbf{0.843} (0.763--0.924) \\

DenseNet169
& \textbf{0.833} (0.799--0.867)
& 0.658 (0.493--0.823)
& 0.828 (0.765--0.891) \\

ResNet18
& 0.769 (0.674--0.864)
& 0.612 (0.349--0.875)
& 0.776 (0.660--0.892) \\

ResNet50
& 0.732 (0.695--0.769)
& 0.452 (0.306--0.599)
& 0.824 (0.726--0.921) \\
\bottomrule
\end{tabular}
\end{adjustbox}
\label{tab:baseline-model-performance}
\vspace{-1cm}
\end{table}




\subsubsection{Fine-Tuning and Training Configuration}

As detailed in Section~\ref{sec:methodology}, we investigated several training configurations to evaluate their impact on tumor classification performance when using TL. Specifically, we assessed the effects of different loss functions, the application of data augmentations, and the integration of a curated dataset.

Table~\ref{tab:experiment-configuration} presents the configurations evaluated using TL, which are the following:
\begin{enumerate*}[label=(\textit{\roman*)}]
    \item The baseline configuration C utilizes only cross-entropy loss without augmentations or the curated dataset.
    \item The configuration CE incorporates the curated dataset into the configuration C.
    \item The configuration CA incorporates the data augmentations to the configuration C.
    \item The configuration CAE includes cross-entropy loss, data augmentations, and curated embeddings.
    \item The configuration F replaces cross-entropy with focal loss \cite{LinFocalDetection} to address class imbalance.
    \item And, the configuration FAE adds data augmentations and curation to the configuration F.
\end{enumerate*}

\begin{table}[t]
\vspace{-.5cm}
\centering
\caption{Experimental configurations across fine-tuning strategies: full fine-tuning (FT), linear probing with fine-tuning (LP-FT), and gradual layer-freezing (G-LF).}
\renewcommand{\arraystretch}{1.2}
\setlength{\tabcolsep}{8pt}
\begin{adjustbox}{width=\linewidth}

\begin{tabular}{@{}ccccc@{}}
\toprule
\textbf{Configuration} & \textbf{Cross-Entropy} & \textbf{Focal Loss} & \textbf{Augmentations} & \textbf{Embeddings} \\
\midrule
C    & \checkmark & -- & -- & -- \\
F    & -- & \checkmark & -- & -- \\
CA   & \checkmark & -- & \checkmark & -- \\
CE   & \checkmark & -- & -- & \checkmark \\
CAE  & \checkmark & -- & \checkmark & \checkmark \\
FAE  & -- & \checkmark & \checkmark & \checkmark \\
\bottomrule
\end{tabular}
\end{adjustbox}
\label{tab:experiment-configuration}
\vspace{-.5cm}
\end{table}

The baseline configuration C demonstrates an improvement in performance over non-pretrained baselines (see Table~\ref{tab:all-performance}). The best result within this configuration was obtained using full FT, yielding an AUC of $0.873 \pm 0.059$. The CE configuration improves this, achieving an AUC of $0.903 \pm 0.037$ using the FT model, indicating that excluding potentially mislabeled healthy samples enhanced the model's generalization capability. In turn, the CA configuration produces substantial performance improvements across all metrics. Under this configuration, the FT model achieved an AUC of $0.950 \pm 0.032$, with a TPR of $0.764$ and a TNR of $0.964$, confirming the well-established effectiveness of augmentation in medical imaging tasks. 

The CAE configuration delivered the highest overall performance across the cross-entropy configuration. The FT model attained an AUC of $0.948 \pm 0.018$, the highest TPR ($0.832$), and a TNR of $0.946$, indicating a well-balanced sensitivity-specificity trade-off and reduced performance variance. Whereas the focal loss in configuration F slightly improved performance over the baseline, its effect was more pronounced in configuration FAE.  In this setting, the FT model achieved an AUC of $0.944 \pm 0.038$, a TPR of $0.874$, and a TNR of $0.876$. Although FAE exhibited slightly higher sensitivity than CAE, CAE maintained superior specificity and lower performance variability, making it the most effective configuration overall.

Integrating data augmentations and embedding-based dataset curation (CAE) resulted in the best overall classification performance. While augmentations alone represent the major improvement in AUC and specificity, adding embeddings in CAE further enhanced sensitivity and consistency. Among the fine-tuning strategies, FT consistently outperformed both LP-FT and G-LF across all configurations. These results highlight the importance of enhancing data diversity and ensuring label quality to optimize tumor classification performance.




\section{Conclusion}
\label{sec:conclusions}

In this paper, we presented a computer-aided framework for the automatic detection of osteosarcoma in CT scans. The proposed system is designed to support physicians, specialists, and radiologists in the diagnostic process, to reduce workload and enhance diagnostic accuracy. To this end, we developed a comprehensive pipeline comprising preprocessing, classification, postprocessing, and visualization stages.

For the classification component, we trained and evaluated a series of convolutional neural network (CNN) models from the DenseNet and ResNet families, employing various transfer learning strategies. Among the evaluated models, DenseNet121 with full fine-tuning and ImageNet pretraining showed the best results, achieving an AUC of 94.8\% and a specificity of 94.6\%.

In addition to classification, the framework includes a visualization module that generates 3D renderings of CT scans, highlighting regions identified as potential tumors. This feature is intended to aid radiological interpretation and support clinical decision-making.

As part of ongoing work, we are integrating segmentation models to extract morphological features such as tumor size and volume, which are relevant for prognosis and treatment planning. For future work, we also plan to investigate ensembles of deep learning models, which have shown promise in related domains such as tuberculosis detection.

Finally, a comprehensive clinical evaluation would help to assess the real-world impact of the proposed framework on medical workflows and to measure its acceptance among healthcare professionals.

\begin{table}
\vspace{-0.5cm}
\centering
\caption{Performance metrics (mean ± 95\% CI) across 5 folds for each experiment configuration and fine-tuning strategy.}
\renewcommand{\arraystretch}{1.4} 
\setlength{\tabcolsep}{8pt} 

\begin{adjustbox}{width=\linewidth}
\begin{tabular}{clccc}
\toprule
\textbf{ID} & \textbf{Strategy} & \textbf{AUC} & \textbf{TPR} & \textbf{TNR} \\
\midrule
C   & FT    & 0.873 (0.814--0.931) & 0.681 (0.651--0.710) & 0.889 (0.859--0.919) \\
C   & LP-FT & 0.846 (0.775--0.916) & 0.705 (0.555--0.856) & 0.858 (0.812--0.903) \\
C   & G-LF  & 0.852 (0.785--0.919) & 0.638 (0.514--0.763) & 0.859 (0.769--0.950) \\
F   & FT    & 0.850 (0.789--0.912) & 0.745 (0.647--0.843) & 0.795 (0.647--0.943) \\
F   & LP-FT & 0.869 (0.824--0.913) & 0.740 (0.662--0.818) & 0.830 (0.722--0.938) \\
F   & G-LF  & 0.853 (0.818--0.888) & 0.711 (0.698--0.723) & 0.842 (0.784--0.900) \\
CA   & FT    & \textbf{0.950 (0.918--0.981)} & 0.764 (0.640--0.888) & \textbf{0.964 (0.953--0.974)} \\
CA   & LP-FT & 0.905 (0.810--1.001) & 0.781 (0.705--0.858) & 0.863 (0.646--1.081) \\
CA   & G-LF  & 0.924 (0.897--0.952) & 0.718 (0.634--0.802) & 0.932 (0.876--0.989) \\
CE   & FT    & 0.903 (0.865--0.940) & 0.688 (0.650--0.726) & 0.911 (0.871--0.952) \\
CE   & LP-FT & 0.849 (0.798--0.900) & 0.667 (0.542--0.792) & 0.857 (0.792--0.922) \\
CE   & G-LF  & 0.830 (0.810--0.851) & 0.617 (0.560--0.674) & 0.872 (0.828--0.917) \\
\textbf{CAE} & \textbf{FT}    & \textbf{0.948 (0.930--0.966)} & \textbf{0.832 (0.805--0.859)} & \textbf{0.946 (0.914--0.977)} \\
CAE & LP-FT & 0.941 (0.914--0.968) & 0.770 (0.714--0.826) & 0.941 (0.904--0.978) \\
CAE & G-LF  & 0.923 (0.883--0.964) & 0.766 (0.694--0.838) & 0.921 (0.857--0.985) \\
FAE & FT    & 0.944 (0.906--0.981) & \textbf{0.874 (0.781--0.967)} & 0.876 (0.834--0.918) \\
FAE & LP-FT & 0.927 (0.906--0.948) & 0.864 (0.813--0.915) & 0.838 (0.729--0.946) \\
FAE & G-LF  & 0.933 (0.917--0.948) & 0.803 (0.716--0.889) & 0.911 (0.854--0.967) \\
\bottomrule
\end{tabular}
\end{adjustbox}
\label{tab:all-performance}
\vspace{-0.5cm}
\end{table}

\section*{Acknowledgments}

This work has been funded by the Spanish Research Agency under project PID2022-138050NB-I00, \textit{``New Scalable I/O Techniques for Hybrid HPC and Data-Intensive Workloads (SCIOT).''}, and by the Santander Bank through a research-oriented master's scholarship (Call AEM\_BS-24/25).
%
%
%
\bibliographystyle{splncs04}
\bibliography{bibliography}

\end{document}